\documentclass[runningheads]{llncs}

\usepackage{eccv}

\usepackage{eccvabbrv}

\usepackage{graphicx}
\usepackage{booktabs}
\usepackage{multirow}
\usepackage{subcaption}
\usepackage{tikz}
\usepackage[accsupp]{axessibility}  % Improves PDF readability for those with disabilities.

\usepackage{hyperref}

\usepackage{orcidlink}

\begin{document}

\title{Causal Supervision of Attention for Affective Behaviour Analysis} 

\titlerunning{Causal Supervision of Attention for Affective Behaviour Analysis}

\author{Nemanja Rašajski\inst{1}\orcidlink{0009-0000-3487-1339}  \and
Konstantinos Makantasis\inst{2}\orcidlink{0000-0002-0889-2766} \and 
Antonios Liapis\inst{1}\orcidlink{0000-0001-5554-1961}
\and
Georgios N. Yannakakis\inst{1}\orcidlink{0000-0001-7793-1450}}

\authorrunning{Rašajski et al.}


\institute{Institute of Digital Games, University of Malta, Malta 
\email{\{nemanja.rasajski,ctriv01,antonios.liapis,georgios.yannakakis\}@um.edu.mt}\\
\and
AI Department, University of Malta, Malta\\
\email{konstantinos.makantasis@um.edu.mt}}

\maketitle

\begin{abstract}
The \textit{11th Affective Behaviour Analysis in-the-wild Competition} includes the Multi-Task Learning Challenge, where participants develop a unified framework for Valence-Arousal Estimation, Expression Recognition, and Action Unit Detection. The challenge lies in learning emotion-related representations that generalize across subjects while remaining robust to spurious factors such as identity, illumination, pose, and demographic variation. To aggregate features extracted by a pre-trained backbone into a compact representation for prediction, attention mechanisms selectively weight the most informative facial regions. However, these attention weights can still capture dataset-specific correlations rather than genuine affective cues. To address this limitation, we propose an attention pooling framework that combines causal supervision with cross-covariance regularization of attention components, encouraging subject-invariant attention and non-redundant representations that improve generalization. Our method achieves $CCC_{VA}=0.5123$ for VA estimation on the official validation set, together with $F_{EX}=0.3116$ and $F_{AU}=0.3974$ for expression recognition and action unit detection, respectively, resulting in an overall $P$ score (the sum of the individual task metrics) of $1.2214$.
\end{abstract}

\section{Introduction}
\label{sec:intro}

Affective computing \cite{picard2000affective} aims to develop intelligent systems capable of perceiving and interpreting human emotions from behavioural signals such as facial expressions, speech, text, and physiological measurements, enabling applications in education \cite{sajjad2023comprehensive}, healthcare \cite{ayata2020emotion, guo2024development}, and robotics \cite{spezialetti2020emotion, marinoiu20183d}. Affective analysis is commonly formulated using three complementary emotion models: the Ekman categorical model \cite{ekman1969repertoire, canal2022survey}, the Russell circumplex model \cite{russell1980circumplex, gunes2010automatic}, and the Facial Action Coding System \cite{ekman1978facial}. To jointly leverage these complementary perspectives, the Affective Behaviour Analysis in-the-Wild Competition introduced the Multi-Task Learning Challenge based on the Aff-Wild2 dataset~\cite{kollias2026affect, kollias2025emotions, kollias2025advancements, kollias2024behaviour4all, kollias20247th, kollias20246th, kollias2024distribution, kollias2023abaw2, kollias2023abaw, kollias2022abaw, kollias2021analysing, kollias2020analysing, kollias2021distribution, kollias2021affect, kollias2019expression, kollias2019face, kollias2019deep, zafeiriou2017aff}, in which models jointly perform valence-arousal (VA) estimation, expression recognition (EX), and action unit (AU) detection while learning shared and task-specific representations.

Despite recent progress, affective behaviour analysis remains challenging because models often learn spurious correlations with subject and environmental factors instead of emotion-relevant cues, limiting generalization to unseen subjects.

\begin{figure}[!t]
     \centering

    % First subfigure (Top)
    \begin{subfigure}{0.8\linewidth}
        \centering
        \includegraphics[width=\linewidth]{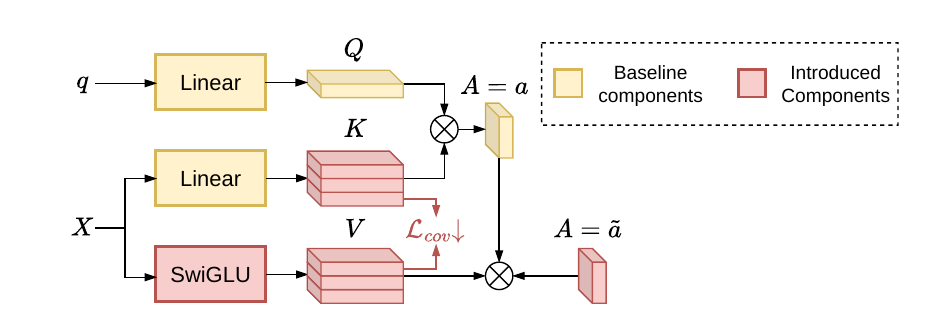}
        \caption{}
        \label{fig:attention_changes}
    \end{subfigure}
    
    \vspace{0.5cm} % Adjust vertical spacing between the subfigures as needed

    % Second subfigure (Bottom)
    \begin{subfigure}{0.95\linewidth}
        \centering
        \includegraphics[width=\linewidth]{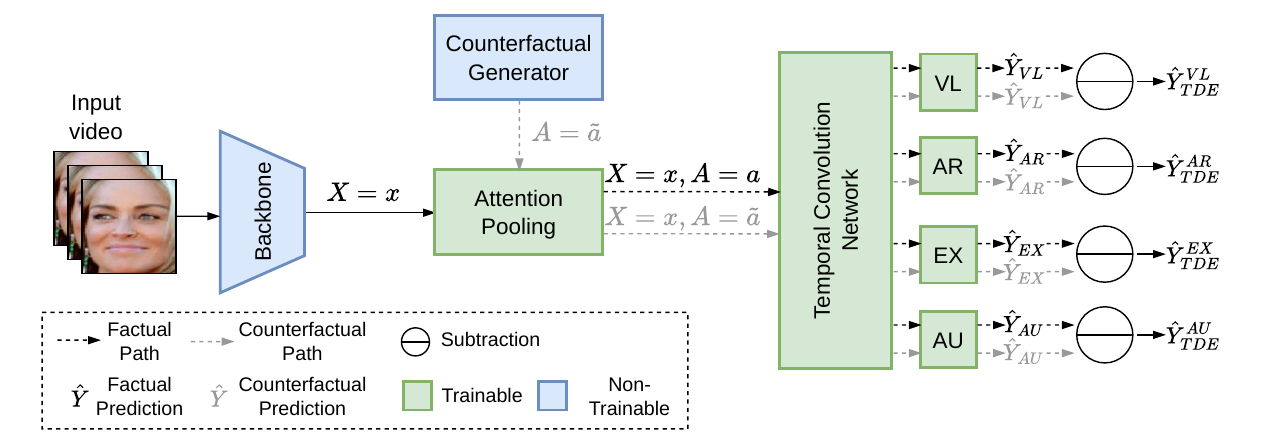}
        \caption{}
        \label{fig:method_pipeline}
    \end{subfigure}

    \caption{Detailed overview of the proposed framework. (a) Our approach extends the baseline with three modifications: counterfactual attention supervision, K-V cross-covariance regularization, and a nonlinear SwiGLU V projection. The proposed components are highlighted in red. (b) Overview of the multi-task affect estimation framework. Feature representations are first extracted using a frozen backbone. Counterfactual samples (dashed grey arrows) are generated and processed alongside factual samples (dashed black arrows) by a temporal encoder, followed by task-specific prediction heads for affect estimation. Finally, the Total Direct Effect is computed for each task.}
    \label{fig:main_figure}
\end{figure}

To improve generalization we propose a causally inspired attention pooling framework for learning robust affective representations. The framework emphasizes emotion-relevant invariant cues while reducing the influence of confounding factors such as identity, pose, and illumination. It incorporates causal supervision to guide attention toward invariant facial regions. It also applies cross-covariance regularization to reduce redundancy between Key (K) representations, which determine attention allocation, and Value (V) representations, which encode the aggregated information. By reducing redundancy between the K and V representations, we enable more effective utilization of their representational capacity, as minimizing shared information allows each representation to encode more complementary and task-relevant features. Finally, a SwiGLU-based nonlinear V projection \cite{shazeer2020glu} further enhances representational capacity (see Fig.~\ref{fig:attention_changes}). Together, these components improve the robustness and expressiveness of emotion representations.

Extensive experiments and ablation studies on the s-Aff-Wild2 dataset demonstrate consistent improvements over standard attention pooling. Under a five-fold cross-evaluation protocol, the proposed method increases the composite score $P$ (defined as the sum of metrics for valence-arousal estimation, expression recognition, and action unit detection) from $0.87$ to $0.95$ with the DINOv2 \cite{oquab2023dinov2} visual encoder and from $1.05$ to $1.19$ with the FRoundation \cite{chettaoui2025froundation} encoder.

\section{Method}
\label{sec:method}

Building on attentive feature aggregation, we introduce three modifications to improve the robustness and representational capacity of affective features: 1) causally motivated supervision to promote attention to facial regions that are both emotion-relevant and invariant to subject identity; 2) cross-covariance independence regularization between the K and V representations to encourage complementary feature encoding; and 3) a SwiGLU-based \cite{shazeer2020glu} nonlinear Value projection to further enhance representational capacity (see Fig.~\ref{fig:attention_changes}).

The framework consists of four stages (see Fig.~\ref{fig:method_pipeline}). First, facial images are transformed into patch embeddings using a frozen pre-trained ViT backbone. Second, patch features are aggregated through cross-attention pooling with a learnable query vector following \cite{bardes2024revisiting}, augmented with the proposed causal supervision and K-V independence regularization. Third, frame-level representations are temporally modeled using a TCN \cite{lea2017temporal}. Finally, task-specific regression and classification heads generate predictions. The following sections describe each component and its associated learning objectives.

\subsection{Causal Supervision for Attention}
\label{subsec:csa}

To improve cross-subject robustness and further emphasize the most important and invariant regions for emotion recognition, we use the Causal Supervision for Attention (CSA) framework \cite{wang2023causal}. This framework introduces an explicit supervision signal based on causal inference. Given patch features $X$, attention weights $A$, and model predictions $\hat{Y}$, CSA models how attention influences the final prediction and measures the contribution of attention by comparing predictions before and after modifying the attention weights. Specifically, the factual prediction is represented as $\hat{Y}_{x,a}=\hat{Y}(X=\mathbf{x},A=\mathbf{a})$, while the counterfactual prediction is obtained by manually replacing the attention weights with counterfactual attention weights, $\hat{Y}_{x,\tilde{a}}=\hat{Y}(X=\mathbf{x},\mathrm{do}(A=\mathbf{\tilde{a}}))$, where the $\mathrm{do}(\cdot)$ operator \cite{pearl2009causality} denotes an intervention that forces the attention to take the value $\mathbf{\tilde{a}}$ rather than the value it would naturally produce. This allows us to measure how the prediction changes when only the attention mechanism is altered (see Fig.~\ref{fig:attention_changes}). The difference between the learned attention weights and the counterfactual attention weights is measured using the Total Direct Effect (TDE) (see Fig.~\ref{fig:method_pipeline}) \cite{vanderweele2016explanation},

\begin{equation}
\hat{Y}_\mathrm{TDE}=\hat{Y}_{x,a}-\hat{Y}_{x,\tilde{a}},
\end{equation}

\noindent which captures the direct causal contribution of attention to the model output. CSA maximizes this effect during training by introducing the following auxiliary supervision objective

\begin{equation}
\mathcal{L}_{CSA}=\mathcal{L}(\hat{Y}_\mathrm{TDE},y)
\label{eq:csa}
\end{equation}

\noindent $y$ is the ground-truth label, and $\mathcal{L}$ is a generic loss function, as this method does not depend on any specific loss formulation.

\textbf{Counterfactual generation.} To generate counterfactual attention vectors, we sample a random vector from a standard Gaussian distribution and normalize it using a softmax operation to obtain a valid attention distribution, where the elements sum to one \cite{wang2023causal}. Formally, $\tilde{\mathbf{a}}=\operatorname{Softmax}(\mathbf{z})$, where $\mathbf{z}\sim\mathcal{N}(\mathbf{0},\mathbf{I})$. The resulting counterfactual attention weights may range from poor (e.g., focusing on background regions) to informative (e.g., emphasizing discriminative facial regions). By maximizing the causal effect across diverse counterfactual attention weights, the model is encouraged to identify causally relevant facial regions consistently, resulting in more robust representations.

\subsection{Cross-Covariance Regularization}
\label{subsec:covarience}

To encourage complementary Key and Value representations in scaled dot-product attention, we introduce a cross-covariance regularization loss (Fig.~\ref{fig:attention_changes}) inspired by \cite{zbontar2021barlow,bardes2021vicreg}. Given the Key and Value representations, $\mathbf{K}$ and $\mathbf{V}$, we first center them along the token dimension,

\begin{equation}
\overline{\mathbf{K}}=\mathbf{K}-\frac{1}{S}\sum_{s=1}^{S}\mathbf{K}_s,
\qquad
\overline{\mathbf{V}}=\mathbf{V}-\frac{1}{S}\sum_{s=1}^{S}\mathbf{V}_s,
\end{equation}

where $S$ is the number of tokens (or spatial locations). The cross-covariance and corresponding regularization loss are then

\begin{equation}
\mathbf{C}_{KV}
=
\frac{1}{S}
\overline{\mathbf{K}}^{\top}
\overline{\mathbf{V}},
\qquad
\mathcal{L}_{\mathrm{cov}}
=
\left\|
\mathbf{C}_{KV}
\right\|_{F}^{2}.
\label{eq:cov}
\end{equation}

Minimizing $\mathcal{L}_{\mathrm{cov}}$ reduces redundancy between the Key and Value feature spaces, encouraging complementary representations.

\subsection{V projection}

We replace the standard fully connected (linear) value projection with a SwiGLU layer~\cite{shazeer2020glu}, increasing its expressiveness so it can encode finer-grained affective cues before attention-based aggregation. Formally, $\mathrm{SwiGLU}(\mathbf{x}) = \mathrm{SiLU}(\mathbf{x}\mathbf{W}_1 + \mathbf{b}_1) \odot (\mathbf{x}\mathbf{W}_2 + \mathbf{b}_2)$, where $\mathbf{W}_1$ and $\mathbf{W}_2$ are learnable weight matrices, $\mathbf{b}_1$ and $\mathbf{b}_2$ are learnable bias vectors, $\odot$ denotes element-wise multiplication, and $\mathrm{SiLU}(z)=z\,\sigma(z)$ is the Sigmoid Linear Unit activation.

\subsection{Training Objectives}
\label{subsec:training_objectives}

We formulate the multi-task objective using task-specific losses. For VA estimation, we optimize a combination of the CCC loss \cite{kollias2018aff} and mean squared error, $\mathcal{L}_{VA}=\mathcal{L}_{CCC}(\hat{y}_{vl},y_{vl})+\mathcal{L}_{MSE}(\hat{y}_{vl},y_{vl})+\mathcal{L}_{CCC}(\hat{y}_{ar},y_{ar})+\mathcal{L}_{MSE}(\hat{y}_{ar},y_{ar})$, where $y_{vl}$ and $y_{ar}$ are the ground-truth valence and arousal labels, and $\hat{y}_{vl}$ and $\hat{y}_{ar}$ are the corresponding predictions. For expression recognition, we use cross-entropy loss, $\mathcal{L}_{EX}=CE(\hat{y}_{ex},y_{ex})$, where $y_{ex}$ and $\hat{y}_{ex}$ represent the ground-truth and predicted expression labels. For AU detection, we employ binary cross-entropy loss, $\mathcal{L}_{AU}=BCE(\hat{y}_{au},y_{au})$, where $y_{au}$ and $\hat{y}_{au}$ denote the ground-truth and predicted AU labels. The joint objective is defined as $\mathcal{L}_{Task}=\mathcal{L}_{AU}+\mathcal{L}_{EX}+\mathcal{L}_{VA}$.

Additionally, we incorporate auxiliary supervision through the CSA loss, defined as $\mathcal{L}_{CSA}=\mathcal{L}_{CSA}^{AU}+\mathcal{L}_{CSA}^{EX}+\mathcal{L}_{CSA}^{VA}$, where each term corresponds to the CSA objective in Eq.~\ref{eq:csa} for the respective task, together with the cross-covariance regularization term from Eq.~\ref{eq:cov}. The final optimization objective is given by $\mathcal{L}_{Overall}=\mathcal{L}_{Task}+\lambda_{CSA}\mathcal{L}_{CSA}+\lambda_{cov}\mathcal{L}_{cov}$. The coefficients $\lambda_{CSA}$ and $\lambda_{cov}$ scale the contributions of the CSA and cross-covariance regularization terms.

\section{Experiments}
\label{sec:experiments}
To evaluate the effectiveness of our proposed framework, we conduct experiments using two backbones pre-trained on different datasets. The first backbone namely FRoundation \cite{chettaoui2025froundation} is specialised for this challenge and is pre-trained on facial images, while the second DINOv2 \cite{oquab2023dinov2} is pre-trained on a general-purpose image corpus. We compare the performance of both backbones against the official baseline, which uses a pre-trained ConvNeXt \cite{liu2022convnet} model with frozen convolutional weights and MixAugment \cite{psaroudakis2022mixaugment} data augmentation, on the official validation set. To further evaluate generalisability, we perform additional experiments using 5-fold cross-validation. We also conduct a comprehensive ablation study to assess the contribution of each component in our framework. The remainder of this section describes the dataset, evaluation metrics, and implementation details.

\subsection{Dataset}
\label{subsec:dataset}
For the MTL track, the organizers provide 142,382 training images and 26,876 validation images from the Aff-Wild2 dataset~\cite{kollias2018aff}. Continuous Valence and Arousal (VA) annotations are provided in the range $[-1, 1]$. Expression (EX) annotations consist of eight categories: Neutral, Anger, Disgust, Fear, Happiness, Sadness, Surprise, and Other. Action Unit (AU) annotations comprise 12 classes: AU1, AU2, AU4, AU6, AU7, AU10, AU12, AU15, AU23, AU24, AU25, and AU26. Following the removal of samples with invalid annotations (AU labels of $-1$, VA labels of $-5$, or EX labels of $-1$), the resulting training and validation sets contain 52,154 and 15,440 images, respectively.

\subsection{Metrics}
\label{subsec:metrics}
The performance measure $P$ as specified by the ABAW MTL challenge is calculated by summing the following components: the mean CCC for valence and arousal, the average $F_1$ Score across all eight expression categories $F_{EX}$, and the average $F_1$ Score across all 12 action units $F_{AU}$. It is mathematically formulated as follows:
\begin{equation}
P = \frac{CCC_{AR} + CCC_{VL}}{2} + F_{EX} + F_{AU}.
\end{equation}

\subsection{Implementation Details}
\label{subsec:implementation_details}
Our method is implemented in PyTorch \cite{paszke2017automatic}, all experiments are performed on a single NVIDIA RTX A6000 GPU using cropped and aligned facial images resized to $224 \times 224$ \cite{kollias2018aff}. For feature extraction, we utilize two models sharing a ViT-S architecture ($16 \times 16$ patches, $256$ tokens, $384$ embedding dimensions): FRoundation \cite{chettaoui2025froundation}, trained on WebFace4M \cite{zhu2021webface260m} with the ArcFace loss \cite{deng2019arcface}, and DINOv2 \cite{oquab2023dinov2}, pre-trained on the general LVD-142M dataset. The TCN module consists of 3 layers with a kernel size of 5. The attention head size is 64, while the number of heads matches the backbone output dimension. Training uses the AdamW optimizer \cite{loshchilov2017decoupled} with a batch size of 8, an initial learning rate of $8\times10^{-5}$, linear warmup for the first 10\% of steps, and a cosine annealing schedule. Models are trained for 15 epochs with early stopping (patience of 5) monitored on a random validation split comprising 15\% of the training data. The coefficients $\lambda_{CSA}$ and $\lambda_{cov}$ are set to $0.1$ and $0.05$ for DINOv2, and to $0.35$ and $0.15$ for FRoundation, respectively.

\section{Results}
\label{sec:results}

\textbf{Comparison with the official baseline and five-fold cross-validation.} Table~\ref{tab:results_cross_validation} presents a comparison with the official baseline and the results of five-fold cross-validation, where Fold~1 corresponds to the official validation split. The baseline achieves an overall challenge score of $P=0.45$ (row~11), while our method substantially improves upon this result, achieving $P=0.96$ (row~1) with the DINOv2 backbone and $P=1.22$ (row~6) with the FRoundation backbone. Across all folds, FRoundation consistently outperforms DINOv2, with the largest gains observed in VA estimation. Expression recognition also benefits consistently, achieving higher $F_{EX}$ across all folds, while action unit detection remains comparable with only minor variations.

\begin{table}[t]
\centering
\caption{The results of the 5-fold cross-validation, where fold 1 corresponds to the official validation set. Final row corresponds to official baseline on the validation set.}
\label{tab:results_cross_validation}
\begin{tabular}{cccccccc}
\hline\hline
Backbone & Fold & $CCC_{VL}$ & $CCC_{AR}$ & $CCC_{VA}$ & $F1_{EX}$ & $F1_{AU}$ & $P$ \\ \hline
\multirow{5}{*}{DINOv2} & 1 & 0.3683 & 0.2663 & 0.3173 & 0.2504 & 0.3982 & 0.9658 \\ \cline{2-8} 
 & 2 & 0.3726 & 0.3466 & 0.3596 & 0.2497 & 0.3849 & 0.9942 \\
 & 3 & 0.2029 & 0.2906 & 0.2468 & 0.1919 & 0.3688 & 0.8074 \\
 & 4 & 0.1748 & 0.4352 & 0.3050 & 0.2162 & 0.3696 & 0.8907 \\
 & 5 & 0.2712 & 0.5597 & 0.4155 & 0.2816 & 0.4294 & 1.1264 \\ \hline
\multirow{5}{*}{FRoundation} & 1 & 0.5299 & 0.4947 & 0.5123 & 0.3116 & 0.3974 & 1.2214 \\ \cline{2-8} 
 & 2 & 0.4726 & 0.3058 & 0.3892 & 0.3044 & 0.3758 & 1.0694 \\
 & 3 & 0.5565 & 0.3952 & 0.4758 & 0.2974 & 0.3989 & 1.1722 \\
 & 4 & 0.5165 & 0.3697 & 0.4431 & 0.3129 & 0.3694 & 1.1254 \\
 & 5 & 0.6008 & 0.6797 & 0.6403 & 0.3063 & 0.4392 & 1.3858 \\ \hline
ConvNeXt & 1 & - & - & - & - & - & 0.4500 \\ \hline\hline
\end{tabular}
\end{table}

\begin{table}[!t]
\centering
\caption{Ablation study of the proposed framework. Results are reported as mean $\pm$ 95\% CI over 5-fold cross-validation, with the highest mean performance in bold. \textbf{SwiGLU}, \textbf{CSA}, and \textbf{Cov} denote the SwiGLU value projection, causal supervision of the attention mechanism, and covariance regularization, respectively, checkmarks ($\checkmark$) indicate enabled components, while crosses ($\times$) indicate that it is disabled. For brevity, only the overall performance metric $P$ is reported.}
\label{tab:results_ablation}
\begin{tabular}{cccccccc}
\hline\hline
Backbone & SwiGLU & CSA & Cov & $P$ \\ \hline
DINOv2 & $\times$ & $\times$ & $\times$ & 0.8730±0.0912 \\
 & $\checkmark$ & $\times$ & $\times$ & 0.8616±0.0972 \\
 & $\checkmark$ & $\checkmark$ & $\times$ & 0.9036±0.1037 \\
 & $\checkmark$ & $\times$ & $\checkmark$ & 0.9317±0.2013 \\
 & $\checkmark$ & $\checkmark$ & $\checkmark$ & \textbf{0.9569±0.1481} \\ \hline
FRoundation & $\times$ & $\times$ & $\times$ & 1.0524±0.0861 \\
 & $\checkmark$ & $\times$ & $\times$ & 1.1265±0.1583 \\
 & $\checkmark$ & $\checkmark$ & $\times$ & 1.1555±0.1004 \\
 & $\checkmark$ & $\times$ & $\checkmark$ & 1.1588±0.1903 \\
 & $\checkmark$ & $\checkmark$ & $\checkmark$ & \textbf{1.1948±0.1498} \\ \hline\hline
\end{tabular}
\end{table}

\textbf{Ablation study.} Table~\ref{tab:results_ablation} reports the ablation results as the mean and 95\% CI of $P$ over five-fold cross-validation (only the composite score is shown). Adding SwiGLU alone has mixed effects, slightly reducing performance with DINOv2 while improving results with FRoundation. However, combining SwiGLU with either causal supervision or covariance regularization consistently improves performance over the baseline. The complete framework achieves the best results for both backbones, with gains of 9.6\% and 13.5\% for DINOv2 and FRoundation, respectively. These findings demonstrate that each component provides complementary benefits, while their combination delivers the most consistent performance improvements.

% \clearpage  % TODO FINAL: This \clearpage needs to be removed from both review and camera-ready versions.

% \section*{Acknowledgements}
% Please insert your acknowledgments here.

\bibliographystyle{splncs04}
\bibliography{main}

@String(NeurIPS = {Adv. Neural Inform. Process. Syst.})

@String(CVPRW = {IEEE Conf. Comput. Vis. Pattern Recog. Worksh.})

@String(AAAI  = {AAAI})

@String(NeurIPS = {NeurIPS})

@String(CVPRW = {CVPRW})

@book{picard2000affective,
  title={Affective computing},
  author={Picard, Rosalind W},
  year={2000},
  publisher={MIT press}
}

@article{kollias2018aff,
  title={Aff-wild2: Extending the aff-wild database for affect recognition},
  author={Kollias, Dimitrios and Zafeiriou, Stefanos},
  journal={arXiv preprint arXiv:1811.07770},
  year={2018}
}

@inproceedings{lea2017temporal,
  title={Temporal convolutional networks for action segmentation and detection},
  author={Lea, Colin and Flynn, Michael D and Vidal, Rene and Reiter, Austin and Hager, Gregory D},
  booktitle={proceedings of the IEEE Conference on Computer Vision and Pattern Recognition},
  pages={156--165},
  year={2017}
}

@article{sajjad2023comprehensive,
  title={A comprehensive survey on deep facial expression recognition: challenges, applications, and future guidelines},
  author={Sajjad, Muhammad and Ullah, Fath U Min and Ullah, Mohib and Christodoulou, Georgia and Cheikh, Faouzi Alaya and Hijji, Mohammad and Muhammad, Khan and Rodrigues, Joel JPC},
  journal={Alexandria Engineering Journal},
  volume={68},
  pages={817--840},
  year={2023},
  publisher={Elsevier}
}

@article{ayata2020emotion,
  title={Emotion recognition from multimodal physiological signals for emotion aware healthcare systems},
  author={Ayata, De{\u{g}}er and Yaslan, Yusuf and Kamasak, Mustafa E},
  journal={Journal of Medical and Biological Engineering},
  volume={40},
  number={2},
  pages={149--157},
  year={2020},
  publisher={Springer}
}

@article{guo2024development,
  title={Development and application of emotion recognition technology—a systematic literature review},
  author={Guo, Runfang and Guo, Hongfei and Wang, Liwen and Chen, Mengmeng and Yang, Dong and Li, Bin},
  journal={BMC psychology},
  volume={12},
  number={1},
  pages={95},
  year={2024},
  publisher={Springer}
}

@article{spezialetti2020emotion,
  title={Emotion recognition for human-robot interaction: Recent advances and future perspectives},
  author={Spezialetti, Matteo and Placidi, Giuseppe and Rossi, Silvia},
  journal={Frontiers in Robotics and AI},
  volume={7},
  pages={532279},
  year={2020},
  publisher={Frontiers Media SA}
}

@inproceedings{marinoiu20183d,
  title={3d human sensing, action and emotion recognition in robot assisted therapy of children with autism},
  author={Marinoiu, Elisabeta and Zanfir, Mihai and Olaru, Vlad and Sminchisescu, Cristian},
  booktitle={Proceedings of the IEEE conference on computer vision and pattern recognition},
  pages={2158--2167},
  year={2018}
}

@book{ekman1969repertoire,
  title={The repertoire of nonverbal behavior: Categories, origins, usage and coding},
  author={Ekman, Paul and Friesen, Wallace V and others},
  volume={1},
  year={1969},
  publisher={Mouton de Gruyter Berlin}
}

@article{canal2022survey,
  title={A survey on facial emotion recognition techniques: A state-of-the-art literature review},
  author={Canal, Felipe Zago and M{\"u}ller, Tobias Rossi and Matias, Jhennifer Cristine and Scotton, Gustavo Gino and de Sa Junior, Antonio Reis and Pozzebon, Eliane and Sobieranski, Antonio Carlos},
  journal={Information Sciences},
  volume={582},
  pages={593--617},
  year={2022},
  publisher={Elsevier}
}

@article{russell1980circumplex,
  title={A circumplex model of affect.},
  author={Russell, James A},
  journal={Journal of personality and social psychology},
  volume={39},
  number={6},
  pages={1161},
  year={1980},
  publisher={American Psychological Association}
}

@article{gunes2010automatic,
  title={Automatic, dimensional and continuous emotion recognition},
  author={Gunes, Hatice and Pantic, Maja},
  journal={International Journal of Synthetic Emotions (IJSE)},
  volume={1},
  number={1},
  pages={68--99},
  year={2010},
  publisher={IGI Global Scientific Publishing}
}

@article{ekman1978facial,
  title={Facial action coding system},
  author={Ekman, Paul and Friesen, Wallace V},
  journal={Environmental Psychology \& Nonverbal Behavior},
  year={1978}
}

@article{chettaoui2025froundation,
  title={Froundation: Are foundation models ready for face recognition?},
  author={Chettaoui, Tahar and Damer, Naser and Boutros, Fadi},
  journal={Image and Vision Computing},
  volume={156},
  pages={105453},
  year={2025},
  publisher={Elsevier}
}

@inproceedings{zhu2021webface260m,
  title={Webface260m: A benchmark unveiling the power of million-scale deep face recognition},
  author={Zhu, Zheng and Huang, Guan and Deng, Jiankang and Ye, Yun and Huang, Junjie and Chen, Xinze and Zhu, Jiagang and Yang, Tian and Lu, Jiwen and Du, Dalong and others},
  booktitle={Proceedings of the IEEE/CVF Conference on Computer Vision and Pattern Recognition},
  pages={10492--10502},
  year={2021}
}

@inproceedings{deng2019arcface,
  title={Arcface: Additive angular margin loss for deep face recognition},
  author={Deng, Jiankang and Guo, Jia and Xue, Niannan and Zafeiriou, Stefanos},
  booktitle={Proceedings of the IEEE/CVF conference on computer vision and pattern recognition},
  pages={4690--4699},
  year={2019}
}

@inproceedings{wang2023causal,
  title={Causal-based supervision of attention in graph neural network: a better and simpler choice towards powerful attention},
  author={Wang, Hongjun and Chen, Jiyuan and Du, Lun and Fu, Qiang and Han, Shi and Song, Xuan},
  booktitle={Proceedings of the Thirty-Second International Joint Conference on Artificial Intelligence},
  year={2023}
}

@article{vanderweele2016explanation,
  title={Explanation in causal inference: developments in mediation and interaction},
  author={VanderWeele, Tyler J},
  journal={International journal of epidemiology},
  volume={45},
  number={6},
  pages={1904--1908},
  year={2016},
  publisher={Oxford University Press}
}

@article{bardes2024revisiting,
  title={Revisiting feature prediction for learning visual representations from video},
  author={Bardes, Adrien and Garrido, Quentin and Ponce, Jean and Chen, Xinlei and Rabbat, Michael and LeCun, Yann and Assran, Mahmoud and Ballas, Nicolas},
  journal={arXiv preprint arXiv:2404.08471},
  year={2024}
}

@inproceedings{
	paszke2017automatic,
	title={Automatic Differentiation in {Pytorch}},
	author={Paszke, Adam and others},
	booktitle={Proc. of {NeurIPS Workshop Autodiff}},
	year={2017}
}

@article{loshchilov2017decoupled,
  title={Decoupled Weight Decay Regularization},
  author={Loshchilov, Ilya and Hutter, Frank},
  journal={arXiv preprint arXiv:1711.05101},
  year={2017}
}

@inproceedings{kollias2026affect,
  title={From Affect to Complex Behavior: Advancing Multimodal Human-Centered AI at the 10th ABAW Workshop \& Competition},
  author={Kollias, Dimitrios and Tzirakis, Panagiotis and Cowen, Alan and Zafeiriou, Stefanos and Kotsia, Irene and Granger, Eric and Pedersoli, Marco and Bacon, Simon and Madsen, Jens and Belharbi, Soufiane and others},
  booktitle={Proceedings of the IEEE/CVF Conference on Computer Vision and Pattern Recognition},
  pages={5302--5311},
  year={2026}
}

@inproceedings{kollias2025emotions, title={From emotions to violence: Multimodal fine-grained behavior analysis at the 9th
abaw}, author={Kollias, Dimitrios and Zafeiriou, Stefanos and Kotsia, Irene and Slabaugh, Greg and Senadeera, Damith Chamalke
and Zheng, Jianian and Yadav, Kaushal Kumar Keshlal and Shao, Chunchang and Hu, Guanyu}, booktitle={Proceedings of the
IEEE/CVF International Conference on Computer Vision}, pages={1--12}, year={2025} }

@inproceedings{kollias2025advancements, title={Advancements in Affective and Behavior Analysis: The 8th ABAW Workshop
and Competition}, author={Kollias, Dimitrios and Tzirakis, Panagiotis and Cowen, Alan and Zafeiriou, Stefanos and Kotsia, Irene
and Granger, Eric and Pedersoli, Marco and Bacon, Simon and Baird, Alice and Gagne, Chris and others}, booktitle={Proceedings
of the Computer Vision and Pattern Recognition Conference}, pages={5572--5583}, year={2025} }

@article{kollias2024behaviour4all, title={Behaviour4all: in-the-wild facial behaviour analysis toolkit}, author={Kollias, Dimitrios
and Shao, Chunchang and Kaloidas, Odysseus and Patras, Ioannis}, journal={arXiv preprint arXiv:2409.17717}, year={2024} }

@inproceedings{kollias20247th, title={7th abaw competition: Multi-task learning and compound expression recognition},
author={Kollias, Dimitrios and Zafeiriou, Stefanos and Kotsia, Irene and Dhall, Abhinav and Ghosh, Shreya and Shao, Chunchang
and Hu, Guanyu}, booktitle={European Conference on Computer Vision}, pages={31--45}, year={2024}, organization={Springer} }

@inproceedings{kollias20246th,title={The 6th affective behavior analysis in-the-wild (abaw) competition},author={Kollias,
Dimitrios and Tzirakis, Panagiotis and Cowen, Alan and Zafeiriou, Stefanos and Kotsia, Irene and Baird, Alice and Gagne, Chris
and Shao, Chunchang and Hu, Guanyu},booktitle={Proceedings of the IEEE/CVF Conference on Computer Vision and Pattern
Recognition},pages={4587--4598},year={2024}}

@inproceedings{kollias2024distribution,title={Distribution matching for multi-task learning of classification tasks: a large-scale
study on faces \& beyond},author={Kollias, Dimitrios and Sharmanska, Viktoriia and Zafeiriou, Stefanos},booktitle={Proceedings
of the AAAI Conference on Artificial Intelligence},volume={38},number={3},pages={2813--2821},year={2024}}

@inproceedings{kollias2023abaw2, title={Abaw: Valence-arousal estimation, expression recognition, action unit detection \&
emotional reaction intensity estimation challenges}, author={Kollias, Dimitrios and Tzirakis, Panagiotis and Baird, Alice and
Cowen, Alan and Zafeiriou, Stefanos}, booktitle={Proceedings of the IEEE/CVF Conference on Computer Vision and Pattern
Recognition}, pages={5888--5897}, year={2023} }

@inproceedings{kollias2023abaw, title={Abaw: Learning from synthetic data \& multi-task learning challenges}, author={Kollias,
Dimitrios}, booktitle={European Conference on Computer Vision}, pages={157--172}, year={2023}, organization={Springer}}

@inproceedings{kollias2022abaw, title={Abaw: Valence-arousal estimation, expression recognition, action unit detection
\& multi-task learning challenges}, author={Kollias, Dimitrios}, booktitle={Proceedings of the IEEE/CVF Conference on
Computer Vision and Pattern Recognition}, pages={2328--2336}, year={2022} }

@inproceedings{kollias2021analysing, title={Analysing affective behavior in the second abaw2 competition},
author={Kollias, Dimitrios and Zafeiriou, Stefanos}, booktitle={Proceedings of the IEEE/CVF International Conference on
Computer Vision}, pages={3652--3660}, year={2021}}

@inproceedings{kollias2020analysing, title={Analysing Affective Behavior in the First ABAW 2020 Competition},
author={Kollias, D and Schulc, A and Hajiyev, E and Zafeiriou, S}, booktitle={2020 15th IEEE International Conference on
Automatic Face and Gesture Recognition (FG 2020)(FG)}, year={2020}, pages={794--800}}

@article{kollias2021distribution, title={Distribution Matching for Heterogeneous Multi-Task Learning: a Large-scale Face
Study}, author={Kollias, Dimitrios and Sharmanska, Viktoriia and Zafeiriou, Stefanos}, journal={arXiv preprint
arXiv:2105.03790}, year={2021} }

@article{kollias2021affect, title={Affect Analysis in-the-wild: Valence-Arousal, Expressions, Action Units and a Unified
Framework}, author={Kollias, Dimitrios and Zafeiriou, Stefanos}, journal={arXiv preprint arXiv:2103.15792}, year={2021}}

@article{kollias2019expression, title={Expression, Affect, Action Unit Recognition: Aff-Wild2, Multi-Task Learning and
ArcFace}, author={Kollias, Dimitrios and Zafeiriou, Stefanos}, journal={arXiv preprint arXiv:1910.04855}, year={2019} }

@article{kollias2019face,title={Face Behavior a la carte: Expressions, Affect and Action Units in a Single Network},
author={Kollias, Dimitrios and Sharmanska, Viktoriia and Zafeiriou, Stefanos}, journal={arXiv preprint arXiv:1910.11111},
year={2019}}

@article{kollias2019deep, title={Deep affect prediction in-the-wild: Aff-wild database and challenge, deep architectures, and
beyond}, author={Kollias, Dimitrios and Tzirakis, Panagiotis and Nicolaou, Mihalis A and Papaioannou, Athanasios and Zhao,
Guoying and Schuller, Bj{\"o}rn and Kotsia, Irene and Zafeiriou, Stefanos}, journal={International Journal of Computer Vision},
pages={1--23}, year={2019}, publisher={Springer} }

@inproceedings{zafeiriou2017aff, title={Aff-wild: Valence and arousal ‘in-the-wild’challenge}, author={Zafeiriou, Stefanos and
Kollias, Dimitrios and Nicolaou, Mihalis A and Papaioannou, Athanasios and Zhao, Guoying and Kotsia, Irene},
booktitle={Computer Vision and Pattern Recognition Workshops (CVPRW), 2017 IEEE Conference on}, pages={1980--1987},
year={2017}, organization={IEEE} }

@article{oquab2023dinov2,
  title={Dinov2: Learning robust visual features without supervision},
  author={Oquab, Maxime and Darcet, Timoth{\'e}e and Moutakanni, Th{\'e}o and Vo, Huy and Szafraniec, Marc and Khalidov, Vasil and Fernandez, Pierre and Haziza, Daniel and Massa, Francisco and El-Nouby, Alaaeldin and others},
  journal={arXiv preprint arXiv:2304.07193},
  year={2023}
}

@article{shazeer2020glu,
  title={Glu variants improve transformer},
  author={Shazeer, Noam},
  journal={arXiv preprint arXiv:2002.05202},
  year={2020}
}

@inproceedings{zbontar2021barlow,
  title={Barlow twins: Self-supervised learning via redundancy reduction},
  author={Zbontar, Jure and Jing, Li and Misra, Ishan and LeCun, Yann and Deny, St{\'e}phane},
  booktitle={International conference on machine learning},
  pages={12310--12320},
  year={2021},
  organization={PMLR}
}

@article{bardes2021vicreg,
  title={Vicreg: Variance-invariance-covariance regularization for self-supervised learning},
  author={Bardes, Adrien and Ponce, Jean and LeCun, Yann},
  journal={arXiv preprint arXiv:2105.04906},
  year={2021}
}

@book{pearl2009causality,
  title={Causality},
  author={Pearl, Judea},
  year={2009},
  publisher={Cambridge university press}
}

@inproceedings{liu2022convnet,
  title={A convnet for the 2020s},
  author={Liu, Zhuang and Mao, Hanzi and Wu, Chao-Yuan and Feichtenhofer, Christoph and Darrell, Trevor and Xie, Saining},
  booktitle={Proceedings of the IEEE/CVF conference on computer vision and pattern recognition},
  pages={11976--11986},
  year={2022}
}

@inproceedings{psaroudakis2022mixaugment,
  title={Mixaugment \& mixup: Augmentation methods for facial expression recognition},
  author={Psaroudakis, Andreas and Kollias, Dimitrios},
  booktitle={Proceedings of the IEEE/CVF Conference on Computer Vision and Pattern Recognition},
  pages={2367--2375},
  year={2022}
}
\end{document}